\pdfoutput=1

\documentclass[11pt]{article}

\usepackage{ACL2023}

\usepackage{times}
\usepackage{placeins}
\usepackage{latexsym}
\usepackage{graphicx}
\usepackage{subcaption}
\usepackage{xcolor}
\usepackage{tikz}
\usepackage{amsmath}
\usepackage{booktabs}
\usepackage[most]{tcolorbox}
\usepackage{listings}
\usepackage{pgfplots}
\usepackage{multirow}
\pgfplotsset{compat=1.17}
\usepackage[utf8]{inputenc} 
\usepackage[T1]{fontenc}    
\usepackage{url}            
\usepackage{booktabs}       
\usepackage{amsfonts}       
\usepackage{nicefrac}       
\usepackage{microtype}      
\usepackage{xcolor}         
\usepackage{amsmath}
\usepackage{amssymb}
\usepackage{amsthm}
\usepackage{bbm}
\usepackage{lipsum}
\usepackage{multirow}
\usepackage{xcolor}
\usepackage{wrapfig}
\usepackage{tabularx}
\usepackage{enumitem}
\usepackage{tikz}
\usepackage{xspace}
\usepackage{adjustbox}

\usepackage[T1]{fontenc}

\usepackage[utf8]{inputenc}

\usepackage{microtype}

\usepackage{inconsolata}

\definecolor{mydarkred}{rgb}{0.6,0,0}
\definecolor{myblue}{HTML}{268BD2}
\definecolor{mygreen}{HTML}{658354}

\newcommand{\DALLE}{DALL$\cdot$E}
\newcommand{\colorswap}{ColorSwap}

\title{ColorSwap: A Color and Word Order Dataset for Multimodal Evaluation}

\author{Jirayu Burapacheep, Ishan Gaur, Agam Bhatia, Tristan Thrush\\
Stanford University\\ \texttt{\{jirayu, tthrush\}@stanford.edu} \\}

\begin{document}

\maketitle
\begin{abstract}
This paper introduces the ColorSwap dataset, designed to assess and improve the proficiency of multimodal models in matching objects with their colors. The dataset is comprised of 2,000 unique image-caption pairs, grouped into 1,000 examples. Each example includes a caption-image pair, along with a ``color-swapped'' pair. We follow the Winoground schema: the two captions in an example have the same words, but the color words have been rearranged to modify different objects. The dataset was created through a novel blend of automated caption and image generation with humans in the loop. We evaluate image-text matching (ITM) and visual language models (VLMs) and find that even the latest ones are still not robust at this task. GPT-4V and LLaVA score 72\% and 42\% on our main VLM metric, although they may improve with more advanced prompting techniques. On the main ITM metric, contrastive models such as CLIP and SigLIP perform close to chance (at 12\% and 30\%, respectively), although the non-contrastive BLIP ITM model is stronger (87\%). We also find that finetuning on fewer than 2,000 examples yields significant performance gains on this out-of-distribution word-order understanding task. The dataset is here: \url{https://github.com/Top34051/colorswap} and here: \url{https://huggingface.co/datasets/stanfordnlp/colorswap}.

\end{abstract}

\section{Introduction}

Recent years have seen remarkable developments in pretrained vision and language models~\cite{radford2021learning, li2022blip, singh2022flava, li2023blip2, li2019visualbert, rombach2021highresolution, betker2023improving, liu2023llava}. Their performance is exceptional in tasks such as visual question-answering~\cite{liu2023llava}, text-to-image generation and manipulation~\cite{minderer2022simple}, and image captioning~\cite{li2022blip, li2023blip2}.

\begin{figure}[t]
\centering
\setlength{\jot}{-2pt}
\includegraphics[width=7.5cm]{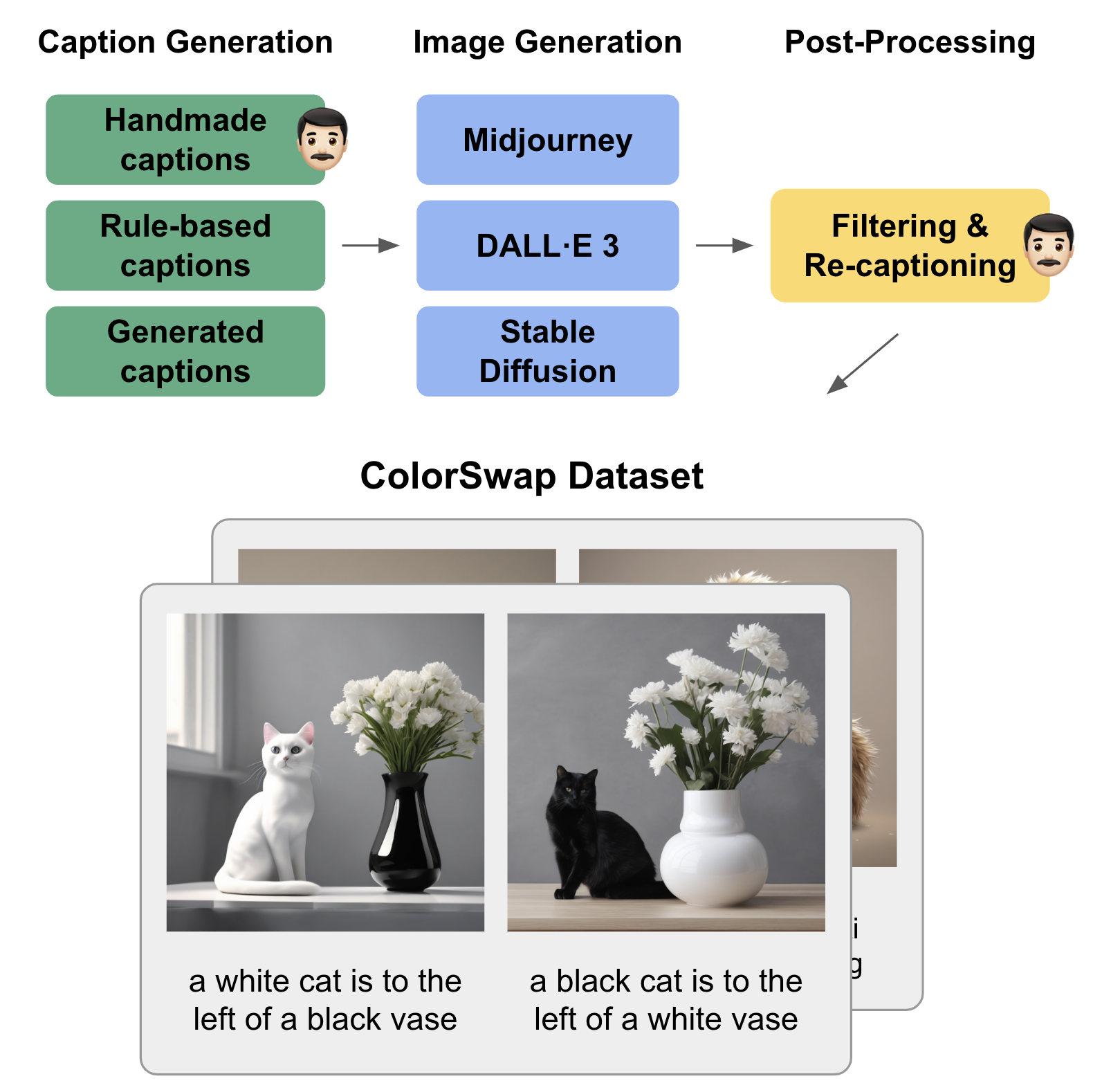}
\caption{An overview of the {\colorswap} dataset creation methodology. The human emoji marks components that require human annotator input.}
\label{fig:colorswap}
\vspace{-0.5cm}
\end{figure}

Despite the success, recent work reveals that vision and language models often struggle to comprehend fine grained distinctions in images~\cite{krojer-etal-2022-image} and compositional relationships, particularly in differentiating captions with the same words but different word orders~\cite{thrush2022winoground}. The Winoground dataset from~\citeauthor{thrush2022winoground} demonstrates that, given two captions composed of the same words in a different order, the performance of well-known models in correctly matching these captions to their respective images is close to random chance. It is not well understood whether known multimodal architectures are even capable, in practice, of learning how to perform well at a balanced word-order-understanding task with small-scale finetuning data. Finetuning has been done on some balanced word-order datasets, but only in settings where half of the images are lacking, as far as we are aware~\cite{yuksekgonul2023when}.

\begin{figure}[ht]
    \centering
    \includegraphics[width=0.6\columnwidth]{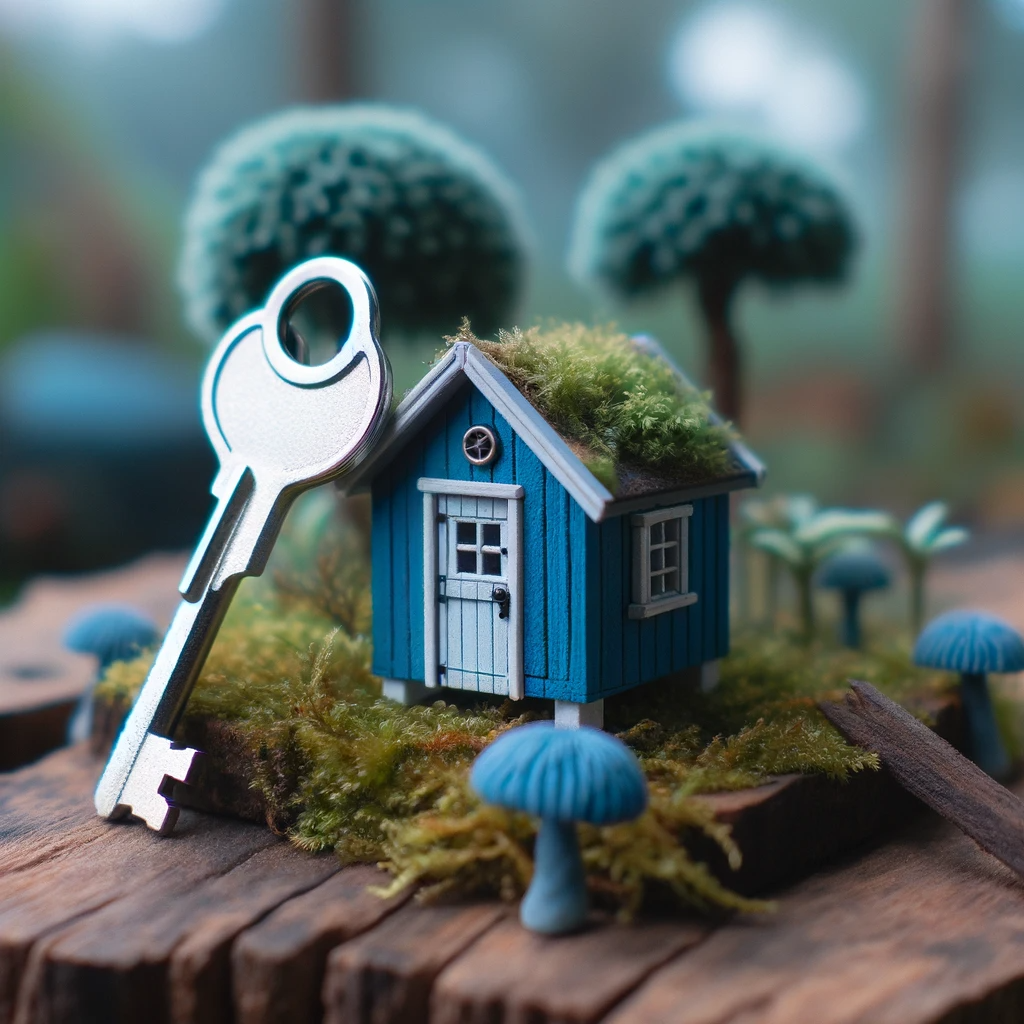}
    \caption{An image from {\DALLE} 3 \cite{betker2023improving} when given the caption ``The key to the shed is blue'' (we ensured that the caption was not rewritten by ChatGPT \cite{chatgpt}). {\DALLE} 3 does not always make this mistake, but it is unreliable. Even though ``the shed is blue'' is a substring, the full sentence is saying that the key is blue. Our dataset does not target difficult cases where colors modify far objects in the string.}
    \label{fig:dalle3fooled}
\end{figure}

To address such questions, we introduce the {\colorswap} dataset, along with a data generation process that enables the quick creation of a larger dataset fitting the same schema as Winoground, as shown in Figure~\ref{fig:colorswap}. Our dataset specifically focuses on a subset of Winoground-style examples, emphasizing the swapping of color words in captions for two main reasons: 1) accurately associating colors with objects is of practical importance in AI-generated art, and 2) it is a conceptually simple and targeted word order understanding task - it is very clear what the human judgements are and why. One of the simplifying features of this task is that color words are often directly adjacent to the objects they modify. For example, ``someone holding a [yellow umbrella] wearing a [white dress]'', although it is important to note that it isn't always this simple: see Figure \ref{fig:dalle3fooled}.


Finally, we conduct evaluations of several vision and language models using this dataset. We find that all models, even GPT-4V \cite{openai2023gpt4}, make a significant number of mistakes. Although, contrastive models such as CLIP and SigLIP struggle drastically. The dataset could be a useful benchmark for contrastive models in particular, but also for visual language models and even industry diffusion models such as Midjourney \cite{midjourney2023}.\footnote{As with other targeted evaluations, there is an implicit confounder: the data is from a different distribution than the training data. ColorSwap is a test of word order understanding on out of distribution data, not just word order understanding.} Further, we found no model which lacks the capability, in practice, to learn some level of generalizable word-order judgements from a fairly small set of naturalistic finetuning data.



\section{ColorSwap}
\label{sec:colorswap}

In this section, we introduce the {\colorswap} dataset, comprised of 1,000 unique examples created by four expert annotators (each with two or more years or NLP experience and living in the United States) with the aid of generative models. Each example includes a caption paired with an image, along with a color-swapped version of the caption and image. We randomly select 700 examples for training and 300 for evaluation.

Our data collection methodology uses three key steps: 1) Caption Generation, where we generate a variety of initial captions using three different methods; 2) Image Generation, which involves employing various text-to-image models to create images corresponding to these captions; and 3) Post-Processing, which includes human review to ensure accuracy, maintain quality, and re-caption images. Post-processing is essential as current text-to-image models often mix up the colors. See Figure~\ref{fig:colorswap} for an illustration of our data collection process and Table~\ref{tab:colorswap} for a summary of the dataset's composition.

\begin{table}[ht]
\centering
\resizebox{0.8\columnwidth}{!} {
\begin{tabular}{lccccc}
    \toprule
    \textbf{Caption} & \textbf{Image} & \textbf{\# Pairs} \\
    \midrule
    Handmade    & Midjourney        & 39 \\
    Handmade    & {\DALLE} 3        & 167 \\
    \midrule
    Rule-based  & Stable Diffusion  & 782 \\
    Rule-based  & {\DALLE} 3        & 394 \\
    \midrule
    Generated   & Stable Diffusion  & 212 \\
    Generated   & {\DALLE} 3        & 406 \\
    \bottomrule
\end{tabular}
}
\caption{\textbf{Number of pairs per method.} Rule-based captions are rewritten by humans during post-processing. Generated captions come from Large Language Models.}
\label{tab:colorswap}
\vspace{-0.5cm}
\end{table}

\subsection{Caption Generation Methods}

\paragraph{Handmade.} In this method, annotators manually create captions by creatively brainstorming scenarios and contexts that involve at least two objects. This process ensures a high degree of originality and diversity in the captions, but is time-consuming. Once we have some captions, we can use them to bootstrap the following two approaches.

\paragraph{Rule-based.} The second method employs a systematic color swapping technique in predefined caption templates, using sets of objects and colors. It generates a broad range of color-object scenarios, though the captions lack creative variability. To tackle this, the post-processing stage involves human review and caption rewriting.

\paragraph{Generative model.} This method leverages generative models, particularly GPT-4~\cite{openai2023gpt4} and Claude-2~\cite{anthropic2023claude}. We prompt these models with examples from our first method and some examples from the Winoground dataset. These models then generate additional caption pairs based on these inputs.

\subsection{Image Generation}

We utilize diffusion models for image generation, a method increasingly used in multimodal dataset creation~\cite{bitton2023breaking, wu2023alsd, lee2023aligning}. For diversity and cost-efficiency, various diffusion models are employed. Despite their limitations in accurately handling the color composition task, we sample multiple images for a given caption and later select the most suitable image. This approach enables us to create ColorSwap even with generative models that would find the task challenging. We use the open-source Stable Diffusion model~\cite{podell2023sdxl}, as well as stronger commercial models such as Midjourney \cite{midjourney2023} and OpenAI's {\DALLE} 3 \cite{betker2023improving}. Details are in Appendix \ref{sec:diffusionmodelusage}.

\subsection{Post-Processing}

\paragraph{Filtering.}

During post-processing, annotators sift through images produced by diffusion models to ensure quality. Their task is to identify the image that most accurately aligns with its caption. If no image meets the criteria, the entire set is discarded. This step is crucial as it ensures that the final dataset is accurate and sensible to human evaluators.

\paragraph{Re-captioning.}

To ensure naturalistic and diverse captions, we do a manual re-captioning of image pairs if they were generated with rule-based captions. See Figure~\ref{fig:recaptioning} for an example. See Appendix \ref{sec:annotatorinterfaces} for snapshots of post-processing interfaces.

\begin{figure}[ht]
    \centering
    \includegraphics[width=0.7\columnwidth]{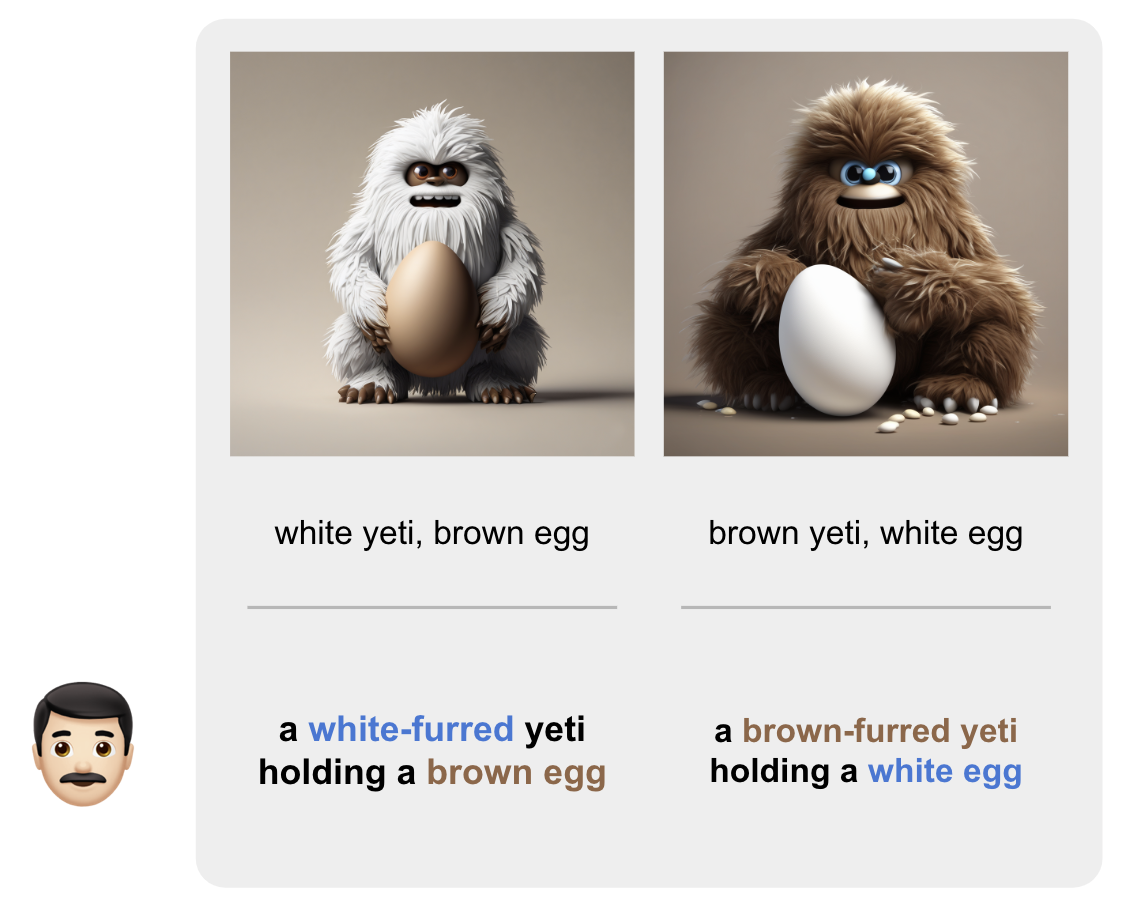}
    \caption{Illustration of re-captioning process.}
    \label{fig:recaptioning}
\vspace{-0.5cm}
\end{figure}

\section{Experiments}


Here, we outline our experiments on the ColorSwap dataset. We evaluate both image-text matching (ITM) models and visual language models. All of these models make a significant number of errors in our color composition task, although contrastive ITM models in particular struggle substantially. Subsequently, we fine-tune these models using the training split of the dataset, aiming to understand whether minimal tuning can significantly improve their ability to understand word order.

\subsection{Evaluation Metrics}
\label{sec:metrics}

To assess model performance, we adopt the three metrics introduced in~\citet{thrush2022winoground}, for Winoground, as our dataset has the same schema. The \textit{text score} measures whether a model can select a correct caption given an image, while the \textit{image score} is about selecting the correct image, given a caption. The \textit{group score} combines both aspects. 

\subsection{Off-the-shelf Models}
\label{sec:experiment-off-the-shelf}

\paragraph{Image-text matching models.}

For ITM models, we evaluate CLIP~\cite{radford2021learning}, FLAVA~\cite{singh2022flava}, BLIP~\cite{li2022blip}, and SigLIP~\cite{zhai2023sigmoid}. FLAVA and BLIP offer two matching methods: 1) a contrastive method, and 2) using cross-modal parameters with an ITM head. CLIP and SigLIP only match text to images in the contrastive way. With these models, we gauage the competnence of these two alternative architectures on the task. Generally, we use the standard base versions of models. More details on model selection are included in Appendix~\ref{sec:experiment-appendix}.

\paragraph{Visual language models.}

We assess LLaVAR~\cite{zhang2023llavar}, LLaVA-1.5~\cite{liu2023improvedllava}, and GPT-4V~\cite{gpt4v}. We follow the VLM Winoground evaluation methodology in~\cite{wu2023role} where we obtain the text score by prompting them to select the correct caption from two options when provided with an image. Similarly, for the image score, we present them with a caption and two images, from which they must select the one that best corresponds to the caption. To avoid positional bias~\cite{zheng2023judging}, we further randomize the order in which captions or images are presented to these models. The scores here are not strictly comparable to those for the ITM models, which consider one image and one caption at a time and output floating point scores. More details on model selection and prompts are included in Appendix~\ref{sec:experiment-appendix}.

\paragraph{Results.}

Table~\ref{tab:off-the-shelf-results} outlines the performance of the models without finetuning. BLIP and SigLIP exhibit superior performance compared to CLIP and FLAVA, both of which are around the levels of random chance. Matching with an ITM head also improves the image-text matching performance, especially for BLIP. Also, GPT-4V, despite its status as a leading closed-source model, still exhibits genuine errors on the simple task posed by this dataset.

In Appendix~\ref{sec:artifacts}, we run an experiment to provide evidence that the models' poor performance is due to the compositional nature of this task and not simply because the images are generated by diffusion models (and so are out of the pretraining distributions). A selection of examples and model responses is included in Appendix~\ref{sec:failure}. We
report confidence intervals for these results in Appendix~\ref{sec:confidence intervals}.

\begin{table}[ht]
\centering
\resizebox{0.85\columnwidth}{!} {
\begin{tabular}{lccccc}
    \toprule
    \textbf{Model \& Method} & \textbf{Text} $\uparrow$ & \textbf{Image} $\uparrow$ & \textbf{Group} $\uparrow$ \\
    \midrule
    \multicolumn{4}{c}{\textbf{Image-text matching models}} \\
    \midrule
    Random chance & 25.00 & 25.00 & 16.67 \\
    \textbf{Contrastive matching} \\
    CLIP        & \textbf{35.67}    & 14.67             & 11.67 \\
    FLAVA       & \textbf{35.33}    & 25.00             & 15.67 \\
    BLIP        & \textbf{75.67}    & \textbf{56.00}    & \textbf{51.00} \\
    SigLIP      & \textbf{61.67}    & \textbf{37.00}    & \textbf{30.33} \\
    \textbf{ITM matching} \\
    FLAVA       & \textbf{36.33}    & 18.67             & 10.33 \\
    BLIP        & \textbf{94.67}    & \textbf{89.00}    & \textbf{87.33} \\
    \midrule
    \multicolumn{4}{c}{\textbf{Visual Language Models}} \\
    \midrule
    Random chance & 25.00 & 25.00 & 6.25 \\
    LLaVAR      & \textbf{27.67}    & \textbf{25.67}    & \textbf{8.33} \\
    LLaVA-1.5   & \textbf{69.67}    & \textbf{54.33}    & \textbf{42.00} \\
    GPT-4V      & \textbf{91.33}    & \textbf{76.33}    & \textbf{72.00} \\
    \bottomrule
\end{tabular}
}
\caption{Performance of models on {\colorswap}. Results above chance are \textbf{bold}. Note that random chance is different in the ITM versus visual language model (VLM) cases because VLMs output a binary value and ITM models output a float (an effectively continuous value).}
\label{tab:off-the-shelf-results}
\vspace{-0.5cm}
\end{table}

\subsection{Fine-tuning on ColorSwap}
\label{sec:experiment-finetuning}


Winoground remains a challenging task, with even advanced models like GPT-4V, using chain-of-thought prompting, struggling to solve it effectively~\cite{wu2023role}. We are not aware of any demonstrations in other papers that provide an answer to whether multimodal models in practice can even be finetuned from a fairly small set of training data to understand any aspects of word order. So, we fine-tune the best performing off-the-shelf BLIP model on our dataset. Due to the continued popularity of CLIP, we also fine-tune the CLIP model. Training details are given in Appendix \ref{sec:tuningdetails}.

\paragraph{Performance improvements post-finetuning.}

For the {\colorswap} dataset, CLIP and BLIP significantly improve on the test set after finetuning on the train set. They are able to learn generalizable knowledge about word order from 1,400 training pairs. In the case of CLIP, performance increases by several times across all metrics. See Table \ref{tab:finetuned-results}.

\begin{table}[ht]
    \centering
    \resizebox{0.8\columnwidth}{!} {
    \begin{tabular}{lccccc}
        \toprule
        \textbf{Model \& Method} & \textbf{Text} $\uparrow$ & \textbf{Image} $\uparrow$ & \textbf{Group} $\uparrow$ \\
        \midrule
        \multicolumn{4}{c}{\textbf{Contrastive matching}} \\
        \midrule
        CLIP        & \textbf{35.67}    & 14.67             & 11.67 \\
        CLIP fine-tuned & \textbf{72.00} & \textbf{69.33} & \textbf{63.00} \\
        BLIP        & \textbf{75.67}    & \textbf{56.00}    & \textbf{51.00} \\
        BLIP fine-tuned & \textbf{86.33} & \textbf{82.67} & \textbf{79.67} \\
        \midrule
        \multicolumn{4}{c}{\textbf{ITM matching}} \\
        \midrule
        BLIP        & \textbf{94.67}    & \textbf{89.00}    & \textbf{87.33} \\
        BLIP fine-tuned & \textbf{96.00} & \textbf{96.67} & \textbf{95.33} \\
        \bottomrule
        \end{tabular}
    }
    \caption{Performance improvements on the {\colorswap} test set post-finetuning on the {\colorswap} train set. Results above chance are \textbf{bold}.}
    \label{tab:finetuned-results}
\vspace{-0.45cm}
\end{table}

Additionally, we extend our evaluation to the Winoground dataset~\cite{thrush2022winoground} and show the results in Table \ref{tab:finetuned-results-winoground}. Even though the finetuned models are able to learn a sensitivity to word order in our minimal color composition task, performance on more complicated compositional tasks remains largely unaffected. Given that there is no practical issue stopping our models from learning sensitivity to word order in the simpler case, compositional understanding for Winoground may simply be a matter of pretraining data or scale.

\begin{table}[ht]
    \centering
    \resizebox{0.78\columnwidth}{!} {
    \begin{tabular}{lccccc}
        \toprule
        \textbf{Model \& Method} & \textbf{Text} $\uparrow$ & \textbf{Image} $\uparrow$ & \textbf{Group} $\uparrow$ \\
        \midrule
        \multicolumn{4}{c}{\textbf{Contrastive matching}} \\
        \midrule
        CLIP            & \textbf{31.25}    & 11.25             & 9.00  \\
        CLIP fine-tuned & 23.00             & 9.75              & 6.25  \\
        BLIP            & \textbf{37.75}    & 15.75             & 12.75 \\
        BLIP fine-tuned & \textbf{33.25}    & 17.50             & 13.25 \\
        \midrule
        \multicolumn{4}{c}{\textbf{ITM matching}} \\
        \midrule
        BLIP            & \textbf{48.50}    & 24.50             & \textbf{20.25} \\
        BLIP fine-tuned & \textbf{46.25}    & \textbf{26.75}    & \textbf{26.75} \\
        \bottomrule
    \end{tabular}
    }
    \caption{Performance on Winoground before and after finetuning on the {\colorswap} dataset. Results above chance are \textbf{bold}. There is no major difference.}
    \label{tab:finetuned-results-winoground}
\vspace{-0.5cm}
\end{table}

\section{Conclusion}

We introduce the {\colorswap} dataset, a collection of 2,000 unique image-caption pairs and 2,000 hard negative pairings. It is specifically designed to evaluate and improve minimal compositional color comprehension abilities of vision and language models. Our methodology for assembling this dataset involved the use of diffusion models for image generation and the incorporation of human input to ensure naturalness and accuracy. We show that popular off-the-shelf vision and language models exhibit extreme limitations in comprehending even this basic color composition task. However, minimal fine-tuning of these models on the {\colorswap} dataset significantly improves their basic understanding of word order.

\section{Limitations}

All of the images in ColorSwap come from diffusion models. The poor performance of models on the dataset could come from the fact that the diffusion images are simply out of distribution, not because these models have issues with color-word compositionality - although we provide some evidence against this in Appendix E. Similarly, the captions may not be ``in-distribution'' for a variety of reasons. There is a risk of misattributing failure reasons. However, if a model scores well, then we can be assured that the model is able to correctly differentiate between color-swapped images in this particular setting. ColorSwap is additionally a static dataset, meaning that models evaluated on it must be vetted for training data contamination.

\section{Ethical Considerations}

For all diffusion models and datasets that we used, we believe that our use is consistent with their intended use and licenses. However, the diffusion models that we used are trained on a variety of opaque sources from the internet, and may make use of data without creators' consent.



\bibliography{anthology,custom}

\begin{thebibliography}{27}
\expandafter\ifx\csname natexlab\endcsname\relax\def\natexlab#1{#1}\fi

\bibitem[{Anthropic(2023)}]{anthropic2023claude}
Anthropic. 2023.
\newblock \url{https://www.anthropic.com/index/ introducing-claude}.

\bibitem[{Betker et~al.(2023)Betker, Goh, Jing, Brooks, Wang, Li, Ouyang, Zhuang, Lee, Guo, Manassra, Dhariwal, Chu, Jiao, and Ramesh}]{betker2023improving}
James Betker, Gabriel Goh, Li~Jing, Tim Brooks, Jianfeng Wang, Linjie Li, Long Ouyang, Juntang Zhuang, Joyce Lee, Yufei Guo, Wesam Manassra, Prafulla Dhariwal, Casey Chu, Yunxin Jiao, and Aditya Ramesh. 2023.
\newblock \href {https://cdn.openai.com/papers/dall-e-3.pdf} {Improving image generation with better captions}.

\bibitem[{Bitton-Guetta et~al.(2023)Bitton-Guetta, Bitton, Hessel, Schmidt, Elovici, Stanovsky, and Schwartz}]{bitton2023breaking}
Nitzan Bitton-Guetta, Yonatan Bitton, Jack Hessel, Ludwig Schmidt, Yuval Elovici, Gabriel Stanovsky, and Roy Schwartz. 2023.
\newblock Breaking common sense: Whoops! a vision-and-language benchmark of synthetic and compositional images.
\newblock In \emph{ICCV}.

\bibitem[{Krojer et~al.(2022)Krojer, Adlakha, Vineet, Goyal, Ponti, and Reddy}]{krojer-etal-2022-image}
Benno Krojer, Vaibhav Adlakha, Vibhav Vineet, Yash Goyal, Edoardo Ponti, and Siva Reddy. 2022.
\newblock \href {https://doi.org/10.18653/v1/2022.acl-long.241} {Image retrieval from contextual descriptions}.
\newblock In \emph{Proceedings of the 60th Annual Meeting of the Association for Computational Linguistics (Volume 1: Long Papers)}, pages 3426--3440, Dublin, Ireland. Association for Computational Linguistics.

\bibitem[{Lee et~al.(2023)Lee, Liu, Ryu, Watkins, Du, Boutilier, Abbeel, Ghavamzadeh, and Gu}]{lee2023aligning}
Kimin Lee, Hao Liu, Moonkyung Ryu, Olivia Watkins, Yuqing Du, Craig Boutilier, Pieter Abbeel, Mohammad Ghavamzadeh, and Shixiang~Shane Gu. 2023.
\newblock \href {http://arxiv.org/abs/2302.12192} {Aligning text-to-image models using human feedback}.

\bibitem[{Li et~al.(2023)Li, Li, Savarese, and Hoi}]{li2023blip2}
Junnan Li, Dongxu Li, Silvio Savarese, and Steven Hoi. 2023.
\newblock \href {http://arxiv.org/abs/2301.12597} {Blip-2: Bootstrapping language-image pre-training with frozen image encoders and large language models}.

\bibitem[{Li et~al.(2022)Li, Li, Xiong, and Hoi}]{li2022blip}
Junnan Li, Dongxu Li, Caiming Xiong, and Steven Hoi. 2022.
\newblock Blip: Bootstrapping language-image pre-training for unified vision-language understanding and generation.
\newblock In \emph{ICML}.

\bibitem[{Li et~al.(2019)Li, Yatskar, Yin, Hsieh, and Chang}]{li2019visualbert}
Liunian~Harold Li, Mark Yatskar, Da~Yin, Cho-Jui Hsieh, and Kai-Wei Chang. 2019.
\newblock \href {http://arxiv.org/abs/1908.03557} {Visualbert: A simple and performant baseline for vision and language}.

\bibitem[{Liu et~al.(2023{\natexlab{a}})Liu, Li, Li, and Lee}]{liu2023improvedllava}
Haotian Liu, Chunyuan Li, Yuheng Li, and Yong~Jae Lee. 2023{\natexlab{a}}.
\newblock Improved baselines with visual instruction tuning.

\bibitem[{Liu et~al.(2023{\natexlab{b}})Liu, Li, Wu, and Lee}]{liu2023llava}
Haotian Liu, Chunyuan Li, Qingyang Wu, and Yong~Jae Lee. 2023{\natexlab{b}}.
\newblock Visual instruction tuning.
\newblock In \emph{NeurIPS}.

\bibitem[{Midjourney(2023)}]{midjourney2023}
Midjourney. 2023.
\newblock \url{https://www.midjourney.com/}.

\bibitem[{Minderer et~al.(2022)Minderer, Gritsenko, Stone, Neumann, Weissenborn, Dosovitskiy, Mahendran, Arnab, Dehghani, Shen, Wang, Zhai, Kipf, and Houlsby}]{minderer2022simple}
Matthias Minderer, Alexey Gritsenko, Austin Stone, Maxim Neumann, Dirk Weissenborn, Alexey Dosovitskiy, Aravindh Mahendran, Anurag Arnab, Mostafa Dehghani, Zhuoran Shen, Xiao Wang, Xiaohua Zhai, Thomas Kipf, and Neil Houlsby. 2022.
\newblock Simple open-vocabulary object detection with vision transformers.
\newblock \emph{ECCV}.

\bibitem[{OpenAI(2022)}]{chatgpt}
OpenAI. 2022.
\newblock \href {https://chat.openai.com/} {Chat{GPT}}.

\bibitem[{OpenAI(2023{\natexlab{a}})}]{openai2023gpt4}
OpenAI. 2023{\natexlab{a}}.
\newblock Gpt-4 technical report.
\newblock \emph{arXiv preprint arXiv:2303.08774}.

\bibitem[{OpenAI(2023{\natexlab{b}})}]{gpt4v}
OpenAI. 2023{\natexlab{b}}.
\newblock Gpt-4v(ision) system card.
\newblock \url{https://openai.com/research/gpt-4v-system-card}.

\bibitem[{Podell et~al.(2023)Podell, English, Lacey, Blattmann, Dockhorn, Müller, Penna, and Rombach}]{podell2023sdxl}
Dustin Podell, Zion English, Kyle Lacey, Andreas Blattmann, Tim Dockhorn, Jonas Müller, Joe Penna, and Robin Rombach. 2023.
\newblock \href {http://arxiv.org/abs/2307.01952} {Sdxl: Improving latent diffusion models for high-resolution image synthesis}.

\bibitem[{Radford et~al.(2021)Radford, Kim, Hallacy, Ramesh, Goh, Agarwal, Sastry, Askell, Mishkin, Clark, Krueger, and Sutskever}]{radford2021learning}
Alec Radford, Jong~Wook Kim, Chris Hallacy, Aditya Ramesh, Gabriel Goh, Sandhini Agarwal, Girish Sastry, Amanda Askell, Pamela Mishkin, Jack Clark, Gretchen Krueger, and Ilya Sutskever. 2021.
\newblock Learning transferable visual models from natural language supervision.
\newblock In \emph{ICML}.

\bibitem[{Rombach et~al.(2021)Rombach, Blattmann, Lorenz, Esser, and Ommer}]{rombach2021highresolution}
Robin Rombach, Andreas Blattmann, Dominik Lorenz, Patrick Esser, and Björn Ommer. 2021.
\newblock \href {http://arxiv.org/abs/2112.10752} {High-resolution image synthesis with latent diffusion models}.

\bibitem[{Rombach et~al.(2022)Rombach, Blattmann, Lorenz, Esser, and Ommer}]{rombach2022highresolution}
Robin Rombach, Andreas Blattmann, Dominik Lorenz, Patrick Esser, and Björn Ommer. 2022.
\newblock \href {http://arxiv.org/abs/2112.10752} {High-resolution image synthesis with latent diffusion models}.

\bibitem[{Singh et~al.(2022)Singh, Hu, Goswami, Couairon, Galuba, Rohrbach, and Kiela}]{singh2022flava}
Amanpreet Singh, Ronghang Hu, Vedanuj Goswami, Guillaume Couairon, Wojciech Galuba, Marcus Rohrbach, and Douwe Kiela. 2022.
\newblock {FLAVA:} {A} foundational language and vision alignment model.
\newblock In \emph{CVPR}.

\bibitem[{Thrush et~al.(2022)Thrush, Jiang, Bartolo, Singh, Kiela, and Ross}]{thrush2022winoground}
Tristan Thrush, Ryan Jiang, Max Bartolo, Amanpreet Singh, Douwe Kiela, and Candace Ross. 2022.
\newblock Winoground: Probing vision and language models for visio-linguistic compositionality.
\newblock \emph{CVPR}.

\bibitem[{Wu et~al.(2023{\natexlab{a}})Wu, Sun, Zhu, Zhao, and Li}]{wu2023alsd}
Xiaoshi Wu, Keqiang Sun, Feng Zhu, Rui Zhao, and Hongsheng Li. 2023{\natexlab{a}}.
\newblock Better aligning text-to-image models with human preference.
\newblock In \emph{ICCV}.

\bibitem[{Wu et~al.(2023{\natexlab{b}})Wu, Zhang, Xiong, Oguz, Gee, and Nie}]{wu2023role}
Yifan Wu, Pengchuan Zhang, Wenhan Xiong, Barlas Oguz, James~C. Gee, and Yixin Nie. 2023{\natexlab{b}}.
\newblock \href {http://arxiv.org/abs/2311.09193} {The role of chain-of-thought in complex vision-language reasoning task}.

\bibitem[{Yuksekgonul et~al.(2023)Yuksekgonul, Bianchi, Kalluri, Jurafsky, and Zou}]{yuksekgonul2023when}
Mert Yuksekgonul, Federico Bianchi, Pratyusha Kalluri, Dan Jurafsky, and James Zou. 2023.
\newblock When and why vision-language models behave like bags-of-words, and what to do about it?
\newblock In \emph{The Eleventh International Conference on Learning Representations}.

\bibitem[{Zhai et~al.(2023)Zhai, Mustafa, Kolesnikov, and Beyer}]{zhai2023sigmoid}
Xiaohua Zhai, Basil Mustafa, Alexander Kolesnikov, and Lucas Beyer. 2023.
\newblock Sigmoid loss for language image pre-training.
\newblock In \emph{ICCV}.

\bibitem[{Zhang et~al.(2023)Zhang, Zhang, Gu, Zhou, Lipka, Yang, and Sun}]{zhang2023llavar}
Yanzhe Zhang, Ruiyi Zhang, Jiuxiang Gu, Yufan Zhou, Nedim Lipka, Diyi Yang, and Tong Sun. 2023.
\newblock \href {http://arxiv.org/abs/2306.17107} {Llavar: Enhanced visual instruction tuning for text-rich image understanding}.

\bibitem[{Zheng et~al.(2023)Zheng, Chiang, Sheng, Zhuang, Wu, Zhuang, Lin, Li, Li, Xing, Zhang, Gonzalez, and Stoica}]{zheng2023judging}
Lianmin Zheng, Wei-Lin Chiang, Ying Sheng, Siyuan Zhuang, Zhanghao Wu, Yonghao Zhuang, Zi~Lin, Zhuohan Li, Dacheng Li, Eric Xing, Hao Zhang, Joseph~E. Gonzalez, and Ion Stoica. 2023.
\newblock Judging {LLM}-as-a-judge with {MT}-bench and chatbot arena.
\newblock In \emph{Conference on Neural Information Processing Systems Datasets and Benchmarks Track}.

\end{thebibliography}
\bibliographystyle{acl_natbib}

\clearpage

\appendix

\section{Diffusion Model Usage Details}
\label{sec:diffusionmodelusage}

\paragraph{Stable Diffusion~\cite{rombach2022highresolution}.} We use Stability AI's Stable Diffusion XL Base 1.0 model for generating images for 994 captions from the rule-based approach. Each caption is suffixed with \texttt{", 4k"} to optimize the image quality, signaling the model to generate high-resolution images. Additionally, we incorporate \texttt{"ugly"} as a negative prompt, which guides the model to avoid outputting deformed images. We set the generation guidance scale to 7.5, the number of denoising steps to 50, the dimension of the generated images to 1,024 $\times$ 1,024 pixels, and the number of images per caption to 8. Although this method is cost-effective, allowing for local running of the model, it sometimes struggles to generate at least one correct image, resulting in a number of failures.

\paragraph{Midjourney~\cite{midjourney2023}.} To tackle harder handmade captions, we explore Midjourney, known for its detailed and artistic image-generation capabilities. We access this model through its Discord server, which offers a straightforward and interactive platform. To ensure the accuracy of the generated images, we often revise and refine the captions. However, this iterative approach resulted in extended processing times, leading to the inclusion of only 39 images from Midjourney in our final dataset. The image dimensions are 512 $\times$ 512 or 1,024 $\times$ 1,024 pixels.

\paragraph{{\DALLE} 3~\cite{betker2023improving}.} Finally, we utilize OpenAI's {\DALLE} 3 for its advanced capabilities in generating highly realistic and detailed images. This model is particularly adept at handling complex and nuanced captions, making it an ideal choice for our more challenging captions from the handmade and generative model approach. While {\DALLE} 3 is preferred for its high performance and scalability through its API, it comes with a high cost of \$0.04 per image. Additionally, {\DALLE} 3 still makes mistakes, which adds to the expense. Our dataset includes a total of 955 images generated using this model. The image dimensions are 1,024 $\times$ 1,024 or 1,024 $\times$ 1,792 pixels.

\section{Annotator Interfaces}
\label{sec:annotatorinterfaces}

In this section, we provide snapshots of the annotator interfaces. See Figure \ref{fig:ui-selection} and Figure \ref{fig:ui-recaption} below.

\section{Experiment Configurations}
\label{sec:experiment-appendix}

\paragraph{Models selection.} For CLIP, we select the base model that utilizes a ViT-B/32 Transformer architecture as an image encoder. For FLAVA, we select the full model that also has a ViT-B/32 as its encoders. For BLIP, we choose their base model trained on COCO dataset. For SigLIP, we select the base model pre-trained on WebLi at resolution 224x224. The Hugging Face model names for each of the models are listed in Table~\ref{tab:models-selection}.

\begin{table}[ht]
    \centering
    \resizebox{0.9\columnwidth}{!} {
    \begin{tabular}{ll}
        \toprule
        \textbf{Model} & \textbf{Hugging Face Model} \\
        \midrule
        CLIP & \href{https://huggingface.co/openai/clip-vit-base-patch32}{\texttt{openai/clip-vit-base-patch32}} \\
        FLAVA & \href{https://huggingface.co/facebook/flava-full}{\texttt{facebook/flava-full}} \\
        BLIP & \href{https://huggingface.co/Salesforce/blip-itm-base-coco}{\texttt{Salesforce/blip-itm-base-coco}} \\
        SigLIP & \href{https://huggingface.co/google/siglip-base-patch16-224}{\texttt{google/siglip-base-patch16-224}} \\
        \midrule
        LLaVAR & \href{https://huggingface.co/truehealth/LLaVar}{\texttt{truehealth/LLaVar}} \\
        LLaVA-1.5 & \href{https://huggingface.co/liuhaotian/llava-v1.5-13b}{\texttt{liuhaotian/llava-v1.5-13b}} \\
        \bottomrule
        \end{tabular}
    }
    \caption{\textbf{Selected models and Hugging Face model names}}
    \label{tab:models-selection}
\end{table}

\begin{figure}[hbt!]
    \centering
    \includegraphics[width=0.9\columnwidth]{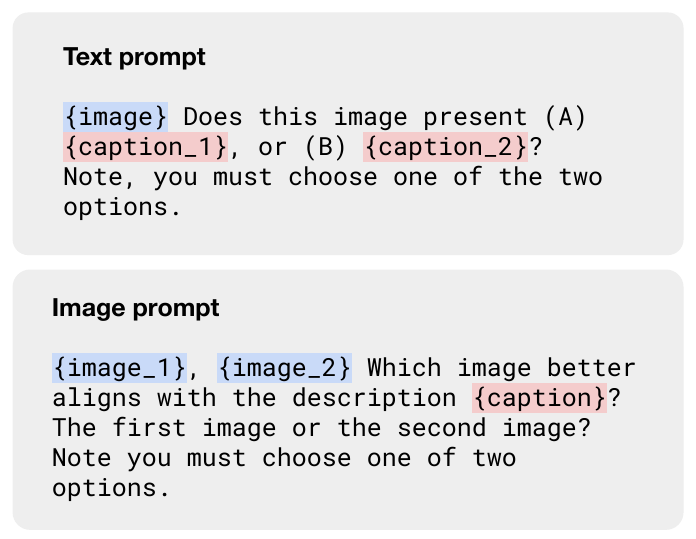}
    \caption{\textbf{Visual language model evaluation prompts.} We replace \textcolor{blue}{\texttt{\{image\}}} with an image and \textcolor{red}{\texttt{\{caption\}}} with an appropriate caption.}
    \label{fig:prompts}
\end{figure}

\paragraph{Visual language model evaluation prompts.} We obtain the text score by prompting visual language models to select the correct caption from two options (Text prompt) and the image score by prompting them to select the correct image from two images (Image prompt). Figure~\ref{fig:prompts} shows these evaluation prompts. For LLaVAR and LLaVA-1.5 where their model interfaces do not directly support multiple images in the input, we horizontally concatenate the images instead.

\section{Finetuning Details}
\label{sec:tuningdetails}

We train CLIP on the training split of the {\colorswap} dataset for 100 epochs. The initial learning rate is $2 \cdot 10^{-5}$ with a linear decay schedule. We employ the Adam optimizer with a weight decay of 0.1 during training, and the batch size is configured to 64. We also fine-tune the BLIP image-text matching model for 100 epochs. The initial learning rate is $1 \cdot 10^{-5}$ with a linear decay schedule, the weight decay is 0.05, and the batch size is 32.

\section{Is Low Performance Caused by OOD Diffusion Images?}
\label{sec:artifacts}

To better understand whether the poor performance on the ColorSwap dataset is due to the images being AI-generated or the models' inability to distinguish compositional color relationships, we run an additional experiment. Instead of choosing between an image-caption pair and a color-swapped version, we assess whether the models can distinguish between an image-caption pair and another randomly selected image-caption pair from our dataset. If the models' poor performance is entirely due to their inability to interpret diffusion-generated images, we would expect them to perform poorly on this task as well. We evaluate some of the worst-performing models from our main experiments, along with BLIP, on this task.

\begin{table}[ht]
\centering
\resizebox{0.8\columnwidth}{!} {
\begin{tabular}{lccccc}
    \toprule
    \textbf{Model \& Method} & \textbf{Text} $\uparrow$ & \textbf{Image} $\uparrow$ & \textbf{Group} $\uparrow$ \\
    \midrule
    Random chance & 25.00 & 25.00 & 16.67 \\
    \textbf{Contrastive matching} \\
    CLIP        & \textbf{100.00}    & \textbf{99.67}    & \textbf{99.67} \\
    FLAVA       & \textbf{99.33}     & \textbf{100.00}   & \textbf{99.33} \\
    BLIP        & \textbf{100.00}    & \textbf{100.00}   & \textbf{100.00} \\
    \textbf{ITM matching} \\
    FLAVA       & \textbf{82.67}     & \textbf{97.67}    & \textbf{81.00} \\
    BLIP        & \textbf{100.00}    & \textbf{100.00}   & \textbf{100.00} \\
    \bottomrule
\end{tabular}
}
\caption{Performance of models on distinguishing matching {\colorswap} image-caption pairs from randomly selected non-matching pairs. Results above chance are \textbf{bold}.}
\label{tab:off-the-shelf-results-artifact}
\end{table}

Table~\ref{tab:off-the-shelf-results-artifact} suggests that the models can distinguish between matching and non-matching image-text pairs even when the images are AI-generated. This provides some evidence that the poor performance on the ColorSwap dataset is not due to a specific issue with the out-of-distribution nature of AI-generated images.

\section{Qualitative GPT-4V Evaluation}
\label{sec:failure}

\begin{figure*}
    \centering
    \includegraphics[width=\linewidth]{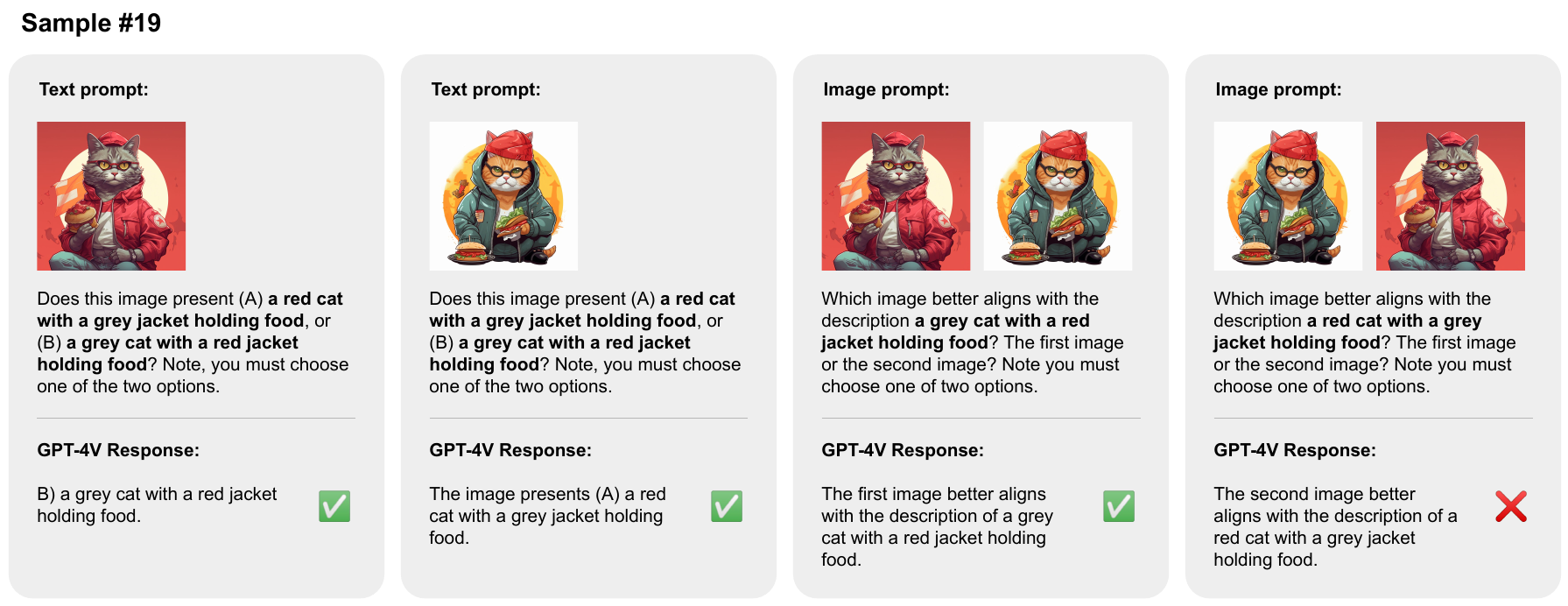}
    \includegraphics[width=\linewidth]{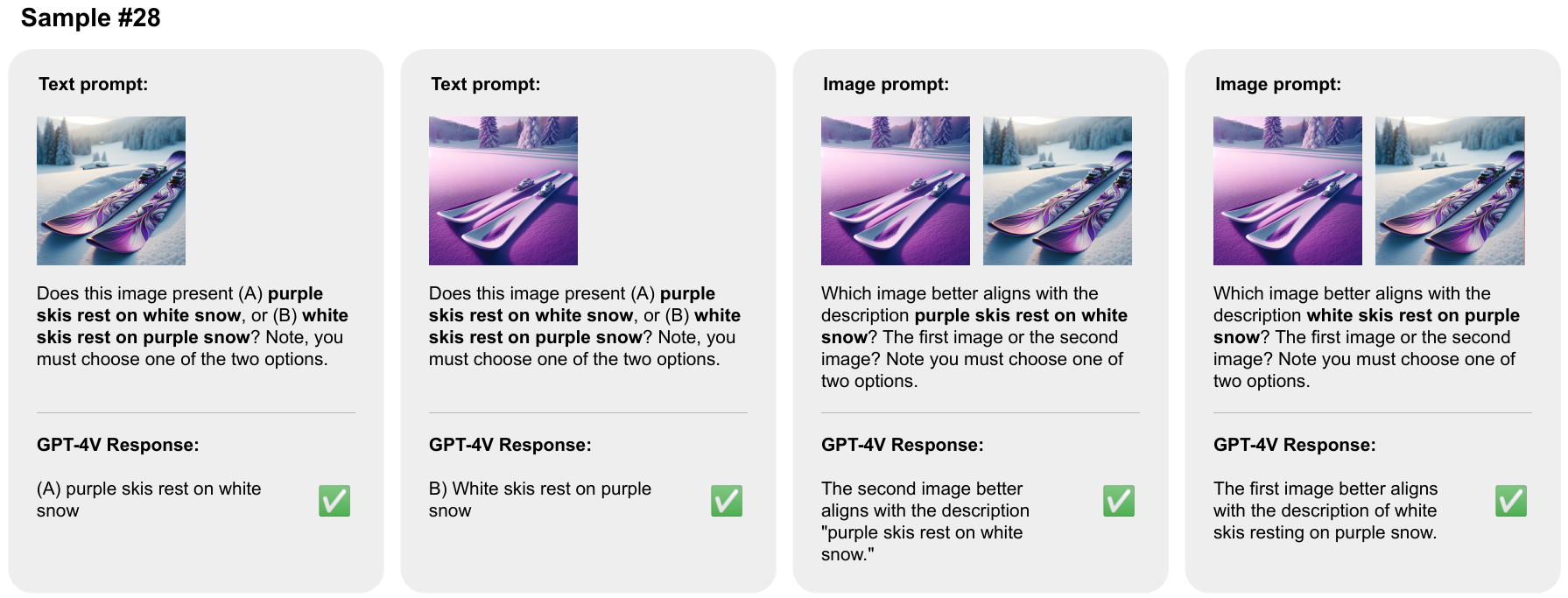}
    \includegraphics[width=\linewidth]{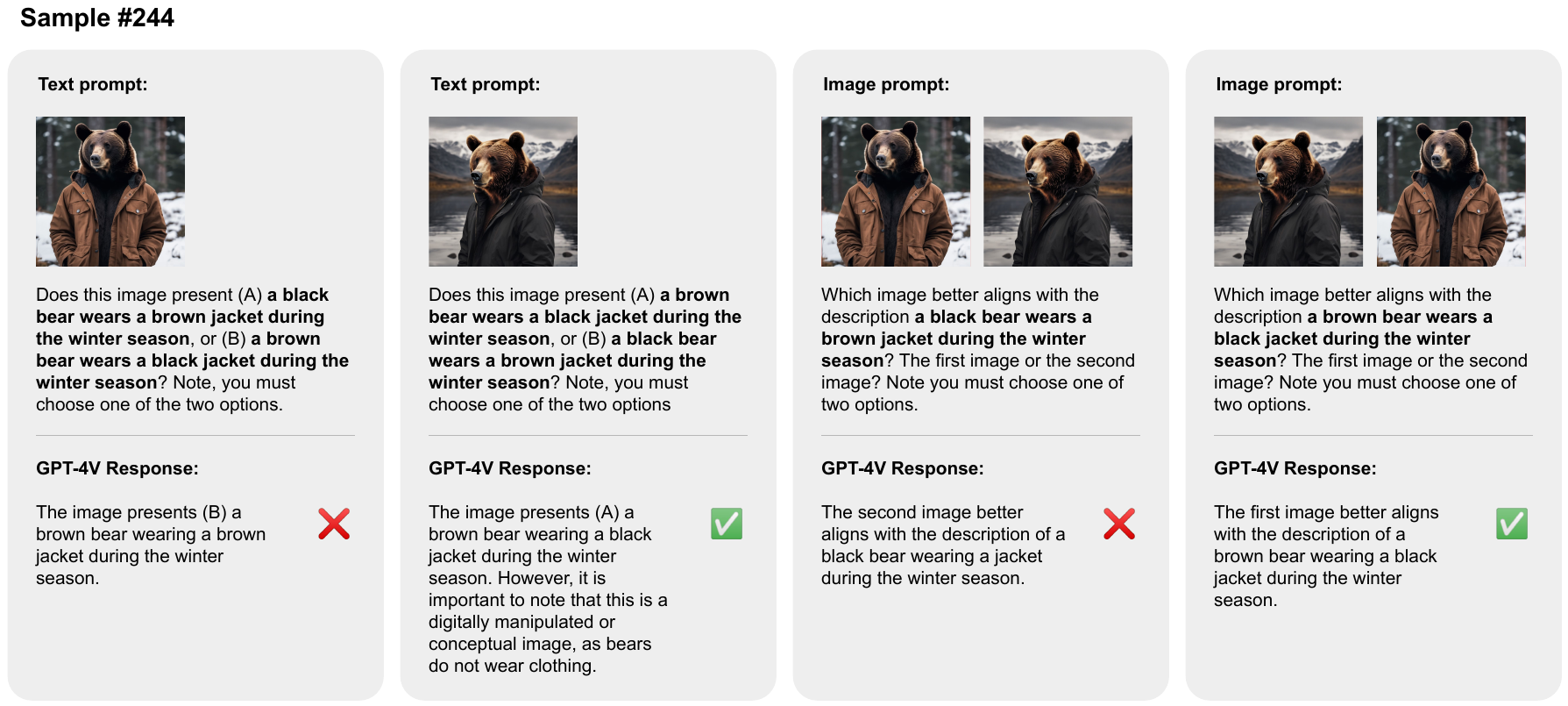}
    \caption{Example 19, 28, and 244 of the ColorSwap dataset. The responses are generated by GPT-4V given different captions and images.}
    \label{fig:examples}
\end{figure*}

In Figure~\ref{fig:examples}, we include three examples from our visual language model evaluation on GPT-4V.

\section{Confidence Intervals}
\label{sec:confidence intervals}

We provide confidence intervals for the overall model results on ColorSwap in Table~\ref{tab:off-the-shelf-results-ci}.

\begin{table*}[ht]
\centering
\resizebox{0.8\linewidth}{!} {
\begin{tabular}{lccccc}
    \toprule
    \textbf{Model \& Method} & \textbf{Text} $\uparrow$ & \textbf{Image} $\uparrow$ & \textbf{Group} $\uparrow$ \\
    \midrule
    \multicolumn{4}{c}{\textbf{Image-text matching models}} \\
    \midrule
    Random chance & 25.00 & 25.00 & 16.67 \\
    \textbf{Contrastive matching} \\
    CLIP        & \textbf{35.67} [\textbf{30.33}, \textbf{41.00}]    & 14.67 [11.33, 19.33]             & 11.67 [8.67, 16.00] \\
    FLAVA       & \textbf{35.33} [\textbf{30.33}, \textbf{41.33}]    & 25.00 [20.33, 30.00]             & 15.67 [11.67, \textbf{20.00}] \\
    BLIP        & \textbf{75.67} [\textbf{70.67}, \textbf{80.33}]    & \textbf{56.00} [\textbf{50.33}, \textbf{61.67}]    & \textbf{51.00} [\textbf{45.33}, \textbf{56.67}] \\
    SigLIP      & \textbf{61.67} [\textbf{56.00}, \textbf{67.00}]    & \textbf{37.00} [\textbf{31.67}, \textbf{42.67}]    & \textbf{30.33} [\textbf{25.33}, \textbf{35.67}] \\
    \textbf{ITM matching} \\
    FLAVA       & \textbf{36.33} [\textbf{31.33}, \textbf{42.00}]    & 18.67 [15.00, 23.67]             & 10.33 [7.67, 14.67] \\
    BLIP        & \textbf{94.67} [\textbf{91.67}, \textbf{97.00}]    & \textbf{89.00} [\textbf{85.00}, \textbf{92.00}]    & \textbf{87.33} [\textbf{83.00}, \textbf{90.67}] \\
    \midrule
    \multicolumn{4}{c}{\textbf{Visual Language Models}} \\
    \midrule
    Random chance & 25.00 & 25.00 & 6.25 \\
    LLaVAR      & \textbf{27.67} [\textbf{22.67}, \textbf{33.00}]    & \textbf{25.67} [\textbf{21.00}, \textbf{31.00}]    & \textbf{8.33} [5.67, \textbf{12.00}] \\
    LLaVA-1.5   & \textbf{69.67} [\textbf{64.33}, \textbf{74.67}]    & \textbf{54.33} [\textbf{48.36}, \textbf{59.67}]    & \textbf{42.00} [\textbf{36.33}, \textbf{47.33}] \\
    GPT-4V      & \textbf{91.33} [\textbf{87.67}, \textbf{94.33}]    & \textbf{76.33} [\textbf{71.33}, \textbf{81.00}]    & \textbf{72.00} [\textbf{66.67}, \textbf{77.00}] \\
    \bottomrule
\end{tabular}
}
\caption{Performance of models on {\colorswap} with confidence intervals. Results above chance are \textbf{bold}. Note that random chance is different in the ITM versus visual language model (VLM) cases because VLMs output a binary value and ITM models output a float (an effectively continuous value).}
\label{tab:off-the-shelf-results-ci}
\end{table*}

\begin{figure*}
    \centering
    \includegraphics[width=0.8\linewidth]{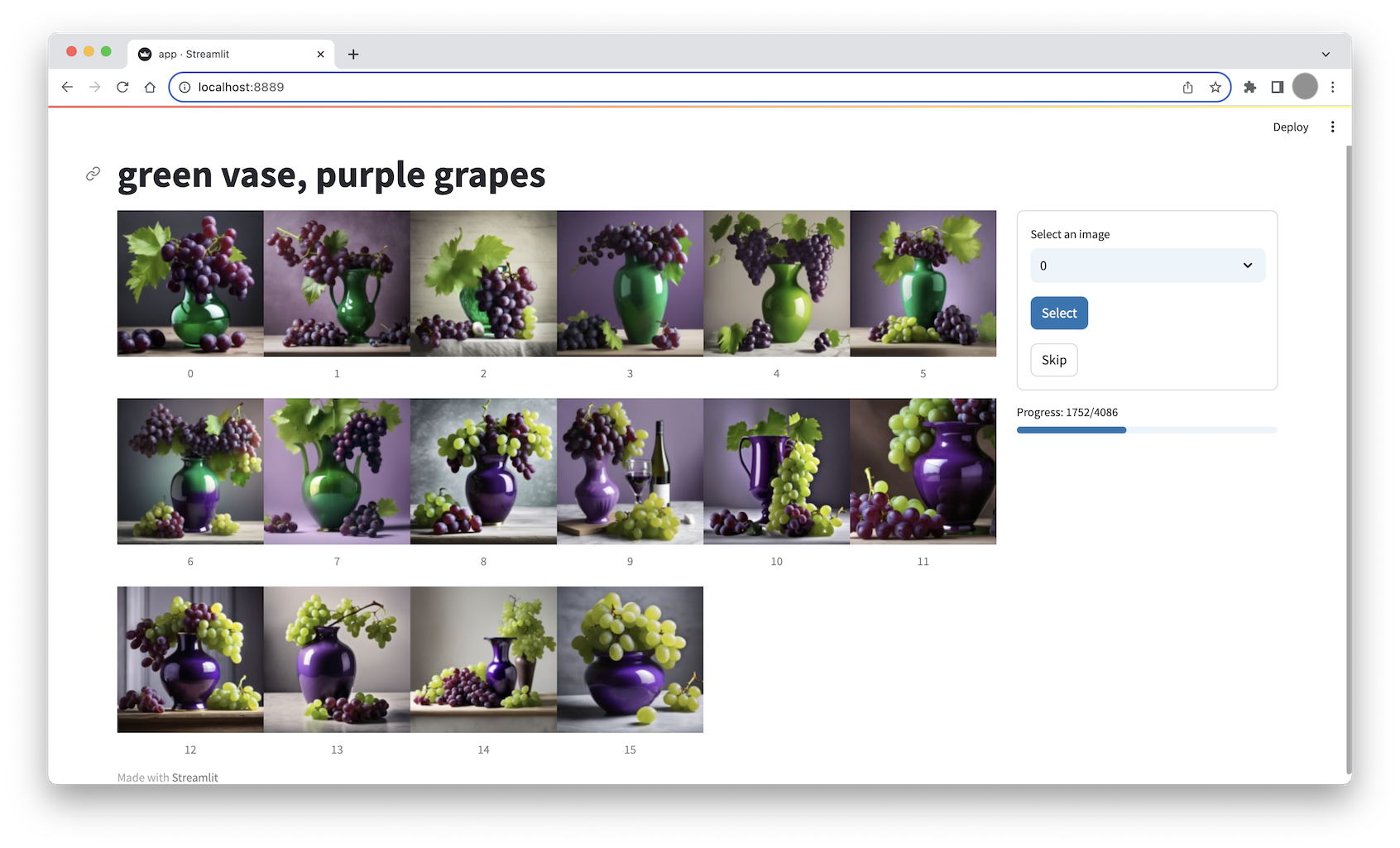}
    \caption{The interface for the selection and filtering process. It allows annotators to choose the image they believe best matches a given caption, with the option to skip if no image seems appropriate. The images presented to the annotators include those generated from the correct and incorrect captions within the same example. This approach is based on the understanding that diffusion models can produce accurate images even from the wrong caption (e.g. a diffusion model could generate an image of a green vase with purple grapes from the caption ``purple vase, green grapes'', which would be correct for the other caption in the pair).}
    \label{fig:ui-selection}
\end{figure*}

\begin{figure*}
    \centering
    \includegraphics[width=0.8\linewidth]{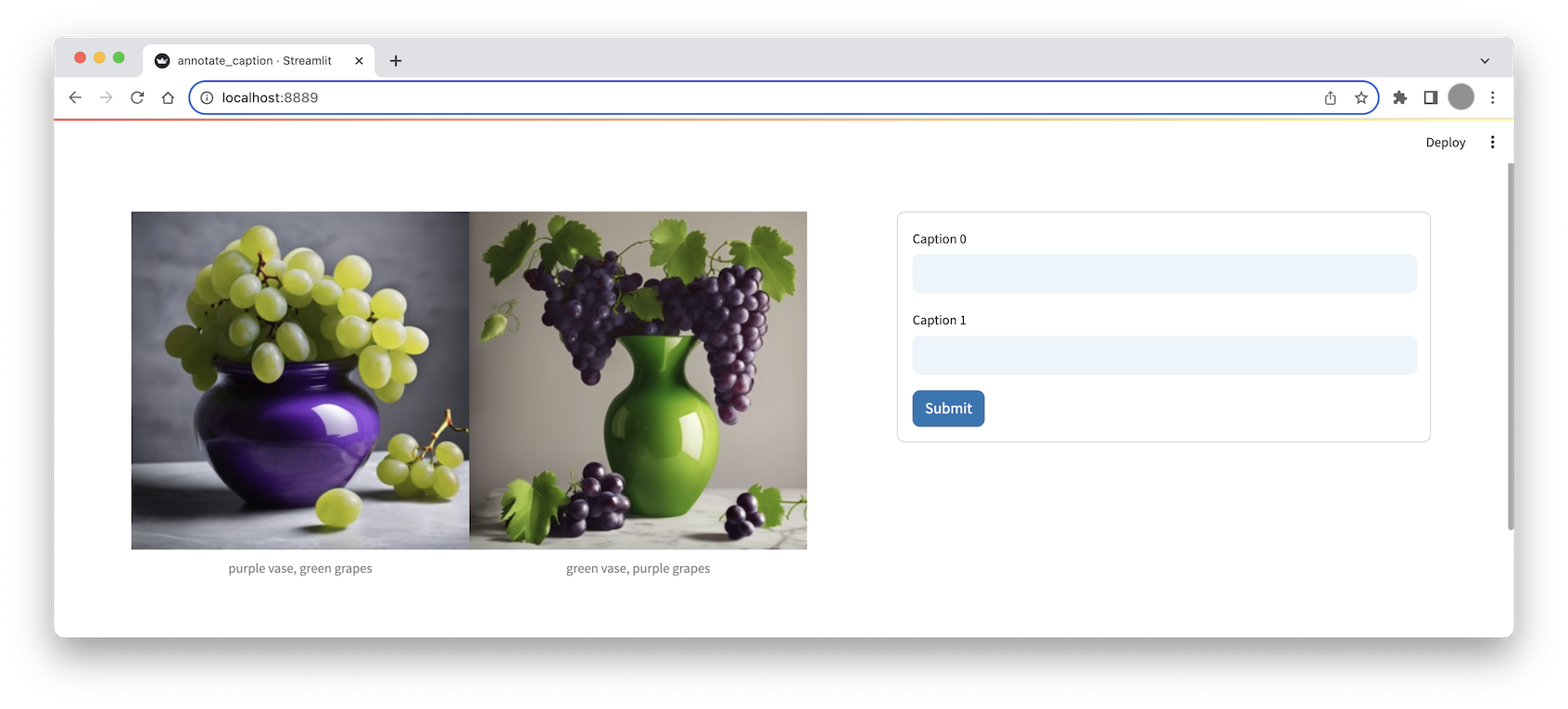}
    \caption{The re-captioning interface. The annotators are provided with two selected images to rewrite the captions. Both images are shown simultaneously so the annotators can infer common things in both pictures and add them to the captions for more nuanced examples.}
    \label{fig:ui-recaption}
\end{figure*}

\end{document}